\documentclass[letterpaper, 10 pt, conference]{ieeeconf}
\usepackage{graphicx} 
\IEEEoverridecommandlockouts
\usepackage{times}
\usepackage{multicol}
\usepackage{tikz}
\usepackage{booktabs}
\usepackage{booktabs} %
\usepackage{pgfplots} %
\pgfplotsset{compat=1.18}
\newcommand{\cell}[2]{#1\,$\pm$\,#2}
\usetikzlibrary{arrows.meta,positioning,fit}
\usepackage[bookmarks=true]{hyperref}

\usepackage{hyperref}
\usepackage{graphicx}		
\usepackage{wrapfig}
\usepackage[format=plain,font=footnotesize,labelfont=bf,labelsep=period]{caption}
\usepackage[export]{adjustbox}
\usepackage{bm}
\usepackage{subcaption}
\usepackage{balance}
\usepackage{dsfont}

\usepackage{amsmath}
\usepackage{mathrsfs}
\usepackage{amssymb}
\usepackage{mathtools}
\usepackage{relsize}
\usepackage{bbm}
\usepackage{cite}
\usepackage{algorithm}
\usepackage{algpseudocode}

\usepackage{pdfsync}
\usepackage[normalem]{ulem}
\usepackage{paralist}	
\usepackage[space]{grffile} 

\usepackage[usenames,dvipsnames]{xcolor}

\usepackage{amsthm}
\usepackage{centernot}

\newtheorem*{theorem*}{Theorem}

\newtheorem*{lemma*}{Lemma}

\newtheorem*{proposition*}{Proposition}

\newcommand{\newsec}[1]{\vspace{0.1cm} \noindent \textbf{#1} }

\definecolor{darkblue}{RGB}{0,0,102}
\definecolor{lightblue}{RGB}{77,77,148}

\definecolor{gold}{RGB}{234, 170, 0}
\definecolor{metallic_gold}{RGB}{139, 111, 78}

\newcommand{\lmat }{\begin{bmatrix}}
\newcommand{\rmat}{\end{bmatrix}}

\newcommand{\Ex}{\mathbb{E}} 
\usepackage{dsfont}
 
\usepackage{svg}
\usepackage{caption}

\makeatletter
\long\def\@makefntext#1{%
  \noindent
  #1}
\makeatother

\IEEEoverridecommandlockouts
\usepackage{amsmath} 
\usepackage{amssymb}  
\usepackage{amsthm}
\usepackage{mathtools}
\usepackage{xcolor}
\usepackage{url}

 \usepackage{tikz} 
\usetikzlibrary{positioning,arrows.meta,shadows.blur,fit}
\usepackage{tikzducks}
\usepackage{mwe} 

\title{\LARGE \bf Architecture Is All You Need: Diversity-Enabled Sweet Spots for Robust Humanoid Locomotion}

\author{*Blake Werner, *Lizhi Yang, Aaron D. Ames \thanks{* denotes equal contribution. All authors affiliated with Caltech MCE.\newline This research is supported in part by the Technology Innovation Institute (TII), BP p.l.c., and by Dow via project \#227027AW.}}

\date{August 2025}

\begin{document}

\maketitle

\begin{abstract}
Robust humanoid locomotion in unstructured environments requires architectures that balance fast low-level stabilization with slower perceptual decision-making. We show that a simple layered control architecture (LCA), a proprioceptive stabilizer running at high rate, coupled with a compact low-rate perceptual policy, enables substantially more robust performance than monolithic end-to-end designs, even when using minimal perception encoders. Through a two-stage training curriculum (blind stabilizer pretraining followed by perceptual fine-tuning), we demonstrate that layered policies consistently outperform one-stage alternatives in both simulation and hardware. On a Unitree G1 humanoid, our approach succeeds across stair and ledge tasks where one-stage perceptual policies fail. These results highlight that architectural separation of timescales, rather than network scale or complexity, is the key enabler for robust perception-conditioned locomotion.
\end{abstract}
\section{Introduction}
Robust humanoid locomotion over mixed and unstructured terrain is a task as old as the platform itself, while still an unsolved problem. 
Sensing of terrain is partial and noisy, contact events are discontinuous, and controllers must react faster than perception can resolve in detail. 
Decades of practice in guidance–navigation–control (GNC) suggest a simple lesson: robustness emerges when fast, low-level stabilization is paired with slower, longer-horizon navigation.
 The canonical example is aerospace GNC \cite{tsien1952automatic,draper1965guidance}: a slow, semantic guidance layer chooses where to go; an intermediate-rate trajectory-generation layer turns goals into feasible references; and a fast feedback control layer tracks those references and rejects disturbances. The same pattern, “slow and flexible” above “fast and rigid,” with well-defined interfaces, recurs across robotics and biological sensorimotor systems \cite{tucker2015control,nakahira2021dess}.
 \begin{figure}[h]
    \centering
    \vspace{.6cm}
 \includegraphics[width=\linewidth]{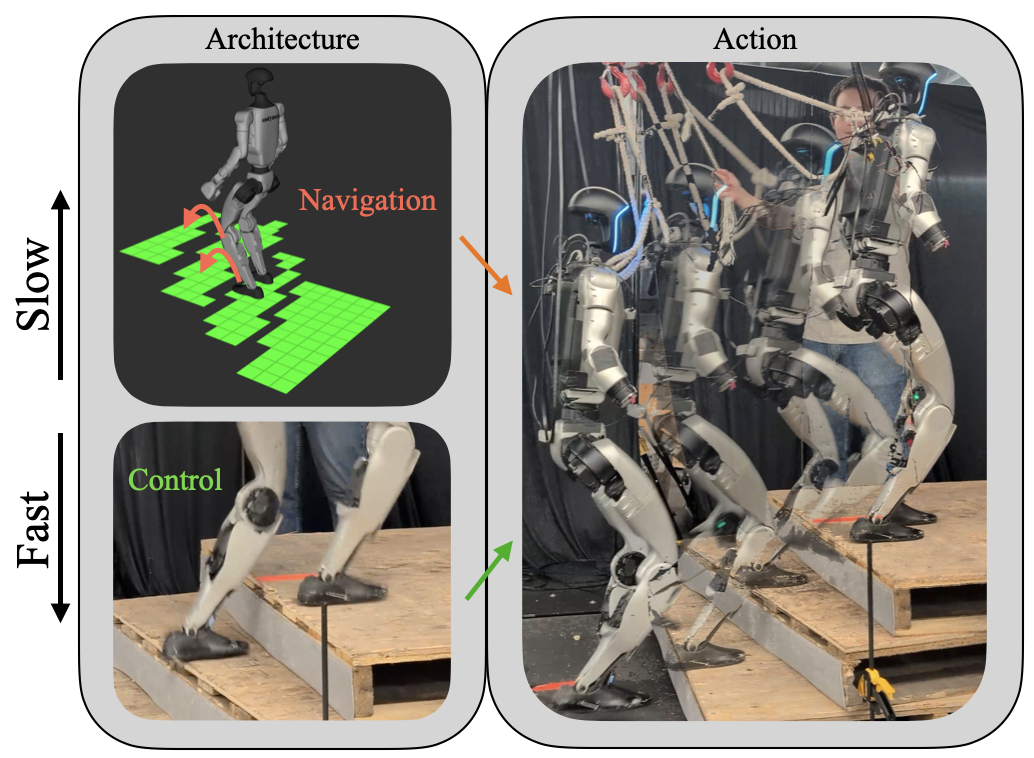}
    \caption{A humanoid robot trained to traverse complex terrain through use of a combination perception information and fast proprioception information. Using this input effectively requires the use of structured architecture in order to produce performant and robust results.}
    \label{fig:toy rewards}
\end{figure}
\begin{figure*}
  \centering
 \includegraphics[width=\linewidth]{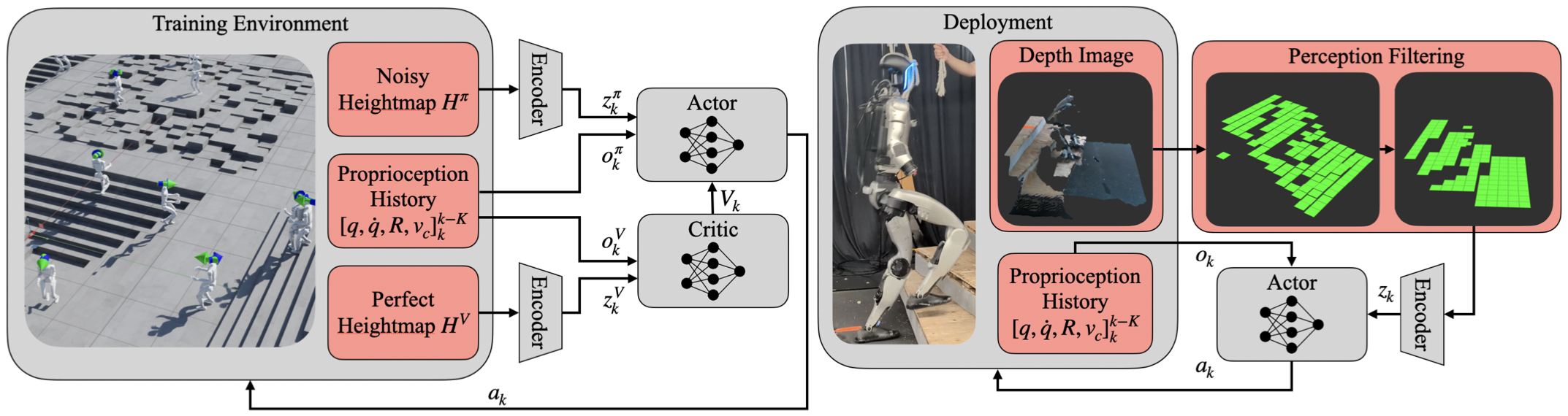}
  \caption{Training and Deployment Overview: both actor and critic are two-stage architectures each with their own perception encoder. The actor receives noisy heightmap information, while the critic receives perfect information, and each receive proprioception history. During deployment, a depth image is filtered and passed through the trained encoder, and the actor combines this with the proprioception history to determine action.}
  \label{fig:dummy}
\end{figure*}

This work takes this observation to its logical extreme and argues that, for high-dimensional perception-conditioned control problems, layered control architecture (LCA) \cite{matni2024lca,full1999templates,rosolia2020multi,rosolia2022unified,csomay2025layered} is the primary driver of robustness. Sophisticated models, learned world representations, or intricate reward shaping help to get the maximum absolute performance, but are not necessary for task success when the stack itself is well-posed.
In a well-posed LCA, information flows through narrow interfaces: references descend (planner $\to$ controller) while tracking error or status ascends (controller $\to$ planner). Crucially, layers operate at different time scales—a design that both reduces computational burden and improves robustness by letting each layer specialize where it is most effective \cite{matni2024lca}. 

The separation of layers in a LCA, together with heterogeneous objectives and information, enables “diversity-enabled sweet spots” (DeSS) \cite{nakahira2021diversity}: the combined stack can outperform any single monolithic component tuned in isolation.
For perception-conditioned humanoid walking, the LCA framing implies a minimal yet sufficient stack: (i) a compact, local-perception navigation encoder that updates at moderate rate to construct an latent space that reflects long-horizon terrain geometry, and (ii) a fast stabilizer that uses proprioception to condition upon this geometry and contend with contact variability. Our method instantiates exactly this two-layer core, with the guidance layer assumed given, aligning with the quantitative architectural principles in \cite{matni2024lca}.

\subsection{Contributions} 

This paper makes two central claims. First, robust locomotion necessarily requires a layered, multi-rate design: a reflexive controller that stabilizes with proprioception at high rate, and a navigation layer that updates more slowly from exteroceptive cues to set short-horizon trajectories. Second, there exists a minimal instantiation of the LCA that can perform complex robust locomotion tasks without the use of heavy machinery: no complex environment estimators, no mixed-integer footstep search, no world models, and no complex network architectures. The performance “sweet spot” arises from different parts of the control architecture taking information at different rates and information budgets rather than from any single sophisticated component.

Concretely, this paper realizes the smallest useful LCA for humanoids: (i) a fast low-level stabilizer (joint-space tracking with largely standard locomotion RL rewards) that runs purely on proprioception, and (ii) a slow navigation policy that consumes a compact local heightmap and allows the low-level to condition itself upon longer-horizon information. Training follows a two-stage curriculum: a blind phase (perception zeroed) that emphasizes stabilization, followed by perception phase that allows for more intelligent longer-horizon planning. This architecture is intentionally plain by design, yet we show it closes the gap to recent methods that rely on richer models or elaborate perception stacks.



Our contributions are as follows:
\begin{itemize}
  \item \textbf{Architecture over complexity} We argue and empirically validate that robust humanoid locomotion \emph{requires} a layered, multi-rate stack; the particular choice of sophisticated models is secondary.
  \item \textbf{Minimal LCA for robust humanoid locomotion} We instantiate a two-layer, two-stage pipeline with standard rewards and a compact local perception interface that performs well on complex locomotion tasks in unstructured terrain.
  \item \textbf{Architecture-isolating ablations} We vary network architectures and training curriculums. Results show that while model details produce only small performance differences, removing the layered structure causes large drops in success and tracking metrics.
\end{itemize}

\subsection{Related Work}

\newsec{Two-stage training pipelines.}
Two-stage curricula appear in several locomotion settings, but for different architectural reasons. On sparse or precarious supports, works emphasize contact selection and balance under limited footholds, effectively prioritizing longer-horizon foot placement behavior before refining stabilization \cite{wang2025beamdojo, zhang2025wococo}. In contrast, other works on challenging terrain follow a “blind-then-vision” strategy: first learn a robust proprioceptive stabilizer, then condition that stabilizer on exteroceptive cues via a slower vision module \cite{duan2024learning, gadde2025no}. Both fall naturally into the LCA view: the first stage trains one layer in isolation (navigation or stabilization), while the second introduces the complementary layer and its interface. Blind stair-traversal and in general rough terrain works \cite{siekmann2021blindstairs,chamorro2024reinforcement, li2025ctbc, singh2024robust, zhang2024bipedal, ji2025robust},  can be seen as extreme instantiations where the navigation/perception layer is absent: such works are often very strong at stabilization but limited in foresight, relying on overfitting to a terrain type from training, and implicitly switching to this 'mode' when encountering this obstacle during deployment. Our design follows the second category, choosing to emphasize proprioceptive stabilization first before adding a conditioning vision module; however, doing so with only a minimal architecture consisting of just those two components. 

\newsec{Perception encoders.}
Across humanoid pipelines, perception does not feed torques directly; instead, visual depth or heightmaps are first encoded, then fused with proprioception downstream \cite{su2024effects}. This preserves rate separation and prevents slow, noisy exteroception from contaminating fast feedback. Examples include perceptive internal models that fuse vision and state estimates for improved foothold selection \cite{long2024learning, he2025attention} and world-model approaches that learn latent representations to inform mid-rate decision making \cite{gu2024advancing}. Our design follows the same pattern but keeps the encoder intentionally compact, simple, and local to maintain a narrow interface between the navigation layer and the stabilizer.

\newsec{Model-based stepping and hybrid stacks.}
Classical “perceptive” footstep planners select contacts via mixed-integer optimization using sensed terrain \cite{acosta2025perceptive}. Hybrid pipelines integrate such planners with model-free RL, letting the planner handle discrete contact choices while RL handles low-level tracking and robustness \cite{su2025lipm}. 
Whole-body methods with sequential contacts and adaptive motion optimization for dexterous humanoids also embody this decomposition: a mid-rate generator proposes feasible references, while a high-rate controller enforces stability and feasibility \cite{liao2025beyondmimic}. In all cases, the pipeline is explicitly layered: planning (navigation) up top, fast feedback below, with narrow reference/feedback channels—precisely the LCA pattern.

\newsec{Student–teacher and distillation.}
Teacher–student pipelines leverage privileged information and rich supervision to train a capable teacher, then distill a deployable student with restricted observations \cite{lee2020learning, kumar2021rma, zhuang2024humanoid,wu2025learn, fan2025one,he2025hover}. From an LCA standpoint, such methods can partially sidestep architectural constraints during training by allowing the teacher to approximate harder, more global solutions before compressing capability into a smaller runtime policy. While highly effective, analyzing their architectural equivalence (e.g., whether the distilled student implicitly embeds a multi-rate decomposition) remains open; we regard this as complementary and leave a deeper treatment for future work.

\section{Methods}

\subsection{Optimization Analysis}

To analyze the complex problem of perception-informed robot control, consider the following optimization:
\begin{align}
    \theta^* = \max_{\theta} \Ex\big [\sum_{k} \gamma^k r(s_k, a_k | \theta) \big ],
\end{align}
where $\theta$ are the network parameters This is the classical one-stage formulation of the reinforcement learning pipeline. Note that while this attempts to solve the global optimal control problem, it suffers from significant sensitivity to initial conditions \cite{picard2021torch} \cite{Goodfellow-et-al-2016}. 

From an optimization perspective, solving the problems in sequence performs a different optimization, with less of the specified sensitivity. Let the parameters of the networks be divided as $\theta = [\theta_x, \theta_y]^T$, with `slow' network parameters $\theta_y$ and `fast' parameters $\theta_x$. Note that in practice, these rates are more frequencies of the signals themselves, rather than the frequency of the controller (as they are all one network running at one speed). By solving the fast-rate optimization first, we solve
\begin{align}
    \theta_x^\dagger = \max_{\theta_x} \Ex\big [\sum_{k} \gamma^k r(s_k, a_k | \theta_x, \theta_{y, 0})\big ],
\end{align}
wherein we maximize the reward conditioned on the fast rate controller parameters $\theta_x$ subject to an initial setting of the slow parameters $\theta_{y, 0}$. In practice, since we will be removing perception from the optimization in stage one of the training, we remove the dependence on $\theta_{y, 0}$, allowing for a simpler optimization more likely to find a satisfactory local maxima. In the second stage of the optimization then, we solve
\begin{align}
        \theta_x^*, \theta_y^* = \max_{\theta_x, \theta_y} \Ex\big [\sum_{k} &\gamma^k r(s_k, a_k | \theta_x, \theta_{y})\big ] \\
        &\mathrm{s.t.} \quad  \theta_{x, 0} = \theta_x^\dagger.
\end{align}
Since we are optimizing over both variables, we perform the same optimization as the one-stage, so in sufficiently regular cases such as strictly concave reward landscapes we are guaranteed the same solution, i.e.  $\theta_{\text{one-stage}}^* = \theta_{\text{two-stage}}^*$. However, in the highly nonconcave reward landscape that we work with, we note that by choosing a good initial condition from the first optimization, we are less susceptible to bad local maxima with large regions of convergence that would otherwise attract the optimization algorithm. 

\begin{figure}[h]
    \centering
    \includegraphics[width=\linewidth]{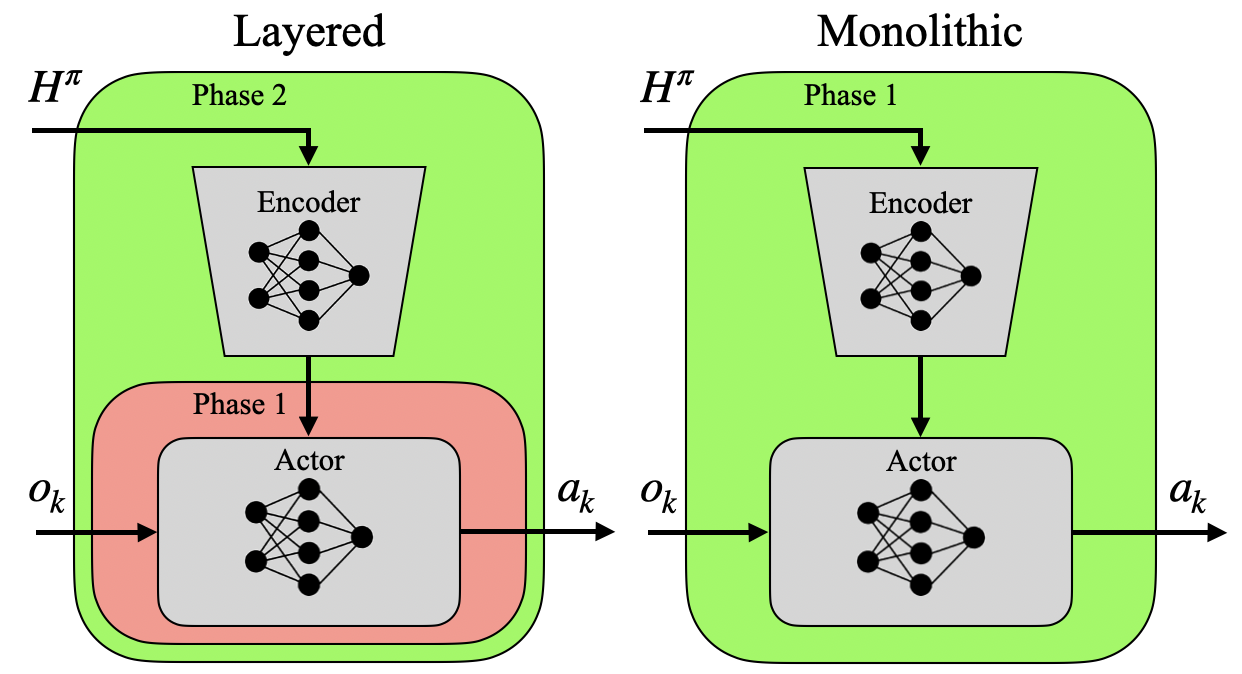}
    \caption{Layered verses monolithic architectures: while the network architecture may be identical, training in two phases allows them to assume the layered control structure.}
    \label{fig:toy rewards}
\end{figure}

\subsection{Observations and Normalization} 
We propose a minimal robust humanoid locomotion pipeline with the goal of illustrating our LCA hypothesis. Let $q$ be joint positions, $\dot q$ joint velocities, $g_b$ the projected gravity direction in body frame, $\omega_b$ the base angular velocity in body frame, $a_{k-1}$ the last applied action, and $u$ the commanded planar velocity. Let $o_k$ be a $K$-length history of robot state information along with current velocity command and last action: $o_k = [q_k, q_{k-1}, ... , q_{k-K},\dot{q}_k, ... \dot{q}_{k-K},\omega_{b,k}, ..., \omega_{b,k-K},  \\g_{b,k}, ... g_{b,k-K}, u, a_{k-1}]$. 

\paragraph{Actor observation} Our actor observation is a concatenation of a the $K$-length history of robot state information and the current velocity command and heightmap: $o^{\pi}_k =[o_k, H^{\pi}]$, where $H^{\pi}\in\mathbb{R}^{11\times 11}$ is a \emph{noisy, sparse} heightmap covering $1.0,\text{m}\times 1.0,\text{m}$ around the robot (robot-centric frame). Note that with a nominal max velocity range of $\pm 0.6$ m/s, 
and the set step period of 0.4s, the robot will take about two steps to move from its current location to that at the edge of the map. Therefore, the map encodes some temporal information (ground height where the robot will be in the future) despite the use of only current heightmap information. Additionally, we do not incorporate perception delays or latency for simplicity and consistency with other methods, but we note that the heightmap signal changes relatively much more slowly than the proprioceptive information. Finally, by using a history of states, we can capture transient and higher-order behaviors than would be allowed by strictly using the current state.
    
\paragraph{Critic observation} To construct our critic observation, we use similar information to the actor with the addition of the world-frame body velocities and a larger and more accurate heightmap with zero noise covering $1.5\text{m}\times 1.5\text{m}$ around the robot. Giving the critic correct ground height information allows for a more accurate advantage function estimate, and the larger heightmap size allows the critic to see further into the `future'. In total, the observation is $o^{V}_k$: $[o_k, v_{\text{base}}, H^V]$. 

Both heightmaps are normalized by subtracting the grid mean (cellwise) and clipping to $[-1,1]$. Zero-centering removes steady-state sim-to-real offsets such as those caused by changed camera mounting and compliance or different motor characteristics and simplifies biases in the MLP during the two-stage training.

\subsection{Network Architecture}

Our network architecture consists of two main components: the perception encoder, a network that takes the perception information and encodes it in a latent representation usable by the main actor network, and the primary actor network that uses a combination of the latent perception information and the standard robot proprioception to determine the robot's actions. Note that in our studies, we consider multiple choices for both the encoder network and the actor network to show the minimal underlying benefits of the actual implementation, instead highlighting the benefit of the layered architecture itself. 

Our choice of encoder is either a small CNN or MLP mapping $H\in\mathbb{R}^{N\times N}$ to an embedding $z_H\in\mathbb{R}^{d_H}$. By ablating this to an MLP, we see if the spatial encoding characteristic of a CNN performs better than a simpler model, even on the small scale of the 11x11 heightmap. A similar network is used to encode the perception information sent to the critic network, the only difference being the larger size of the input. 

The actor network, where we consider both the LSTM and MLP network architectures takes a concatenation of proprioception and perception information and outputs actions $a_t$ as position setpoints, which are then tracked by joint-level PD controllers. 

\subsection{Rewards}
We construct a number of rewards designed for our specific task to guide training, allowing for feasible performance on all of the proposed architectures, where we keep these rewards consistent. We use the following notation: we notate feet $i\in\{L,R\}$, the contact indicator as $\mathcal{C}_i(k)=\mathbf{1}\!\big(\max_j\|\mathbf{F}_i^{(j)}(k)\|_2>F_\star\big)$ with $F_\star{=}1$\,N, the planar foot velocity as $\mathbf{v}^{xy}_i(k)$, foot pitch as $\theta_{i,\text{pitch}}(t)$, foot height as $z_i(k)$, and phase as $\phi_i(k)$.

\paragraph{Phase–contact consistency reward}
For $\tau{=}0.55$, $\varepsilon{=}5{\times}10^{-3}$, let the stance intent be
\[
s_i(t) = \mathbf{1}\!\big(\,\phi_i(k)<\tau \ \lor\ \|\mathbf{u}_{\text{cmd}}(k)\|_2<\varepsilon\,\big),
\; .
\]
Detected contact is $c_i(t)=\mathcal{C}_i(k)$. The reward is the XNOR agreement:
\begin{align*}
r_{\text{phase}}(k) \;=\; &  \sum_{i\in\{L,R\}} \mathbf{1}\big(c_i(k)=s_i(k)\big) \\
\;=\; &  2 - \sum_{i\in\{L,R\}}\big|c_i(k)-s_i(k)\big|.
\end{align*}
\textit{Standing case.} When $\|\mathbf{u}_{\text{cmd}}(t)\|_2<\varepsilon$, we have $s_L=s_R=1$, so $r_{\text{phase}}$ rewards double support and acts as the standing reward. We therefore do not include a separate standing term.

\paragraph{Foot–strike cost}
Penalize lateral ground–reaction forces (scuffs, edge kicks):
\[
r_{\text{strike}} \;=\; \sum_{i\in\{L,R\}} \big\| \mathbf{F}^{xy}_i \big\|_2,
\qquad
\mathbf{F}^{xy}_i=\begin{bmatrix}F^x_{i}&F^y_{i}\end{bmatrix}.
\]

\paragraph{Feet sliding cost}
Suppress planar slip during stance:
\[
r_{\text{slide}} \;=\; \sum_{i\in\{L,R\}} \mathcal{C}_i(k)\,\big\|\mathbf{v}^{xy}_i(k)\big\|_2^{2}.
\]

\paragraph{Feet orientation (flatness) cost}
Encourage flat feet in contact with smooth saturation:
\[
r_{\text{orient}} \;=\; 1 - \exp\!\Big(-k_\theta \sum_{i\in\{L,R\}} \mathcal{C}_i(k)\,\big|\theta_{i,\text{pitch}}(k)\big|\Big),
\quad k_\theta{=}25.
\]

\paragraph{Feet clearance (swing height) cost}
Penalize deviation from target swing height only when \emph{not} in contact:
\[
r_{\text{clear}} \;=\; \sum_{i\in\{L,R\}} \big(1-\mathcal{C}_i(k)\big)\;
\Big(\tfrac{z_i(k)-h_i^\star(k)}{h_{\text{scale}}}\Big)^{2}\;
g_i(k),
\]
where $h_i^\star(k)$ is the nominal swing height (collapsing to foot thickness near zero command), $g_i(k)=\tanh\!\big(\kappa\|\mathbf{v}^{xy}_i(k)\|\big)$ gates by step activity, and $h_{\text{scale}}$ normalizes units.

\paragraph{Total reward} We combine the terms with positive weights and subtract the penalties from the standard locomotion rewards: \[ r=r_{\text{locomotion}} + 0.5 \; r_{\text{phase}} - \; r_{\text{strike}}- 0.2 \; r_{\text{slide}} - \; r_{\text{orient}} + \; r_{\text{clear}}. \]

\subsection{Two-Stage Curriculum} 
Our training curriculum consists of two stages, wherein the robot first learns to traverse complex terrains without perception information, yielding a good baseline along with stabilization capabilities in order to deal with unseen obstacles or perturbations, then is given heightfield information, allowing the robot to learn longer-horizon behavior. 

\paragraph{Stage 1 (blind stabilization)} We set $H\equiv 0$ for the actor (though the critic is still given full information), and train in an environment made up of a quarter respectively of up-stairs, down-stairs, uneven terrain, and flat terrain tasks. 

\paragraph{Stage 2 (perception-critical)} Re-enable $H$ for the actor. This allows the robot to make longer-horizon plans based on the local terrain, or put differently, condition the blind policy on the perceived surroundings. 

\subsection{Perception Filtering}

During deployment, we perform a few stages of filtering for our perception stack in order to curate our data in a way that is usable by the policy. Note that while other works have used more complex perception filtering pipelines such as U-Nets and Transformers \cite{he2025attention} \cite{duan2024learning}, ours is intentionally simple, robust, and requires minimal tuning. Our input is a noisy, dense, depth image from the concatenation of two depth images. We then perform the following steps.

\paragraph{Downsampling}
We aggregate the dense pointcloud to an $11\times 11$ grid over $1.0,\text{m}\times 1.0,\text{m}$ by taking the minimum of each of the valid point heights in each cell. While this method does make the downsampling more sensitive to noise, the minimum value approximation allows for a more correct height estimate in situations such as stair occlusions, where the higher stair occludes the lower one, but the occluded values should be mapped to the lower stair height. 

\paragraph{Outlier Rejection} 
We then compute the mean $\mu_z$ and standard deviation $\sigma_z$ of the grid heights, then clamp outliers to the mean value. Here, we consider outliers to be cells $(g_x, g_y, g_z)$ such that
\begin{align}
    |g_z - \mu_z| \geq \gamma \sigma_z
\end{align}
where $\gamma$ is a tuned parameter. While the prior step deals with disturbances and outliers at the point level, this helps to deal with outliers at the grid cell level, such as the robot's own legs and small anomalies in the terrain. 

\paragraph{Zero-mean and Quantize}
We then subtract $\mu_z$ from all the heightmap values in order to zero-center them and clip each to $[-1,1]$ as the observation requires. Finally, we quantize to buckets corresponding to `steps' of 5cm. This is the smallest ground height perturbation in simulation, and the stabilization of the policy seems robust to smaller perturbations.

\begin{table*}[t]
  \centering
  \small
  \caption{Simulation results across two tasks. Metrics: success rate ($\uparrow$), contacts per step ($\downarrow$), tracking error in cm ($\downarrow$).}
  \label{tab:sim_results_two_tasks}
  \resizebox{\textwidth}{!}{%
  \begin{tabular}{@{}l ccc ccc@{}}
    \toprule
    & \multicolumn{3}{c}{\textbf{Stairs}}
    & \multicolumn{3}{c}{\textbf{Uneven Terrain}} \\
    \cmidrule(lr){2-4}\cmidrule(lr){5-7}
    \textbf{Policy}
      & $R_{\mathrm{succ}}$ (\%, $\uparrow$) & Contacts/step (\%, $\downarrow$) & Track (cm/s, $\downarrow$)
      & $R_{\mathrm{succ}}$ (\%, $\uparrow$) & Contacts/step (\%, $\downarrow$) & Track (cm/s, $\downarrow$) \\
    \midrule
    \multicolumn{7}{l}{\textit{Medium difficulty}} \\
    Blind            & 98.30 & \cell{0.85}{0.20} & \cell{13.40}{1.43} & 97.86 & \cell{2.28}{0.34} & \cell{13.90}{1.44} \\
    One-Stage MLP    & 99.00 & \cell{0.64}{0.16} & \cell{14.10}{1.39} & 98.00 & \cell{1.66}{0.31} & \cell{15.30}{1.45} \\
    Two-Stage MLP    & 97.90 & \cell{1.04}{0.18} & \cell{13.30}{1.42} & 98.60 & \cell{0.80}{0.15} & \cell{13.40}{1.35} \\
    One-Stage LSTM   & 97.40 & \cell{0.55}{0.20} & \cell{13.10}{1.36} & 98.80 & \cell{0.54}{0.15} & \cell{13.30}{1.37} \\
    Two-Stage LSTM   & 99.40 & \cell{0.26}{0.11} & \cell{13.50}{1.29} & 99.40 & \cell{0.34}{0.12} & \cell{14.40}{1.32} \\
    One-Stage CNN    & 98.90 & \cell{0.99}{0.23} & \cell{13.50}{1.49} & 97.80 & \cell{2.74}{0.38} & \cell{14.80}{1.54} \\
    Two-Stage CNN    & 98.00 & \cell{0.79}{0.26} & \cell{13.60}{1.34} & 98.50 & \cell{2.79}{0.39} & \cell{15.70}{1.54} \\
    \addlinespace[2pt]
    \multicolumn{7}{l}{\textit{Hard difficulty}} \\
    Blind            & 93.08 & \cell{0.81}{0.20} & \cell{15.10}{1.56} & 64.63 & \cell{2.79}{0.43} & \cell{17.70}{2.05} \\
    One-Stage MLP    & 92.80 & \cell{1.59}{0.45} & \cell{16.10}{1.60} & 61.60 & \cell{3.36}{0.54} & \cell{18.20}{1.92} \\
    \bf{Two-Stage MLP}    & \bf{90.48} & \cell{\bf{0.56}}{0.14} & \cell{\bf{13.40}}{1.41} & \bf{70.96} & \cell{\bf{3.10}}{0.55} & \cell{\bf{17.70}}{2.00} \\
    One-Stage LSTM   & 85.33 & \cell{3.24}{0.47} & \cell{17.20}{1.96} & 58.02 & \cell{3.10}{0.52} & \cell{18.10}{2.01} \\
    \bf{Two-Stage LSTM}   & \bf{95.01} & \cell{\bf{0.78}}{0.25} & \cell{\bf{16.10}}{1.64} & \bf{72.36} & \cell{\bf{3.06}}{0.51} & \cell{\bf{17.60}}{1.92} \\
    One-Stage CNN    & 96.10 & \cell{2.15}{0.39} & \cell{15.30}{1.68} & 63.93 & \cell{5.45}{0.69} & \cell{19.10}{2.18} \\
    \bf{Two-Stage CNN}    & \bf{98.40} & \cell{\bf{0.83}}{0.20} & \cell{\bf{14.47}}{1.42} & \bf{71.95} & \cell{\bf{5.68}}{0.86} & \cell{\bf{19.10}}{2.12} \\
    \bottomrule
  \end{tabular}
  }
\end{table*}

\section{Simulation Experiments}

\subsection{Training}

\begin{figure}[h]
    \centering
    \includegraphics[width=0.9\linewidth]{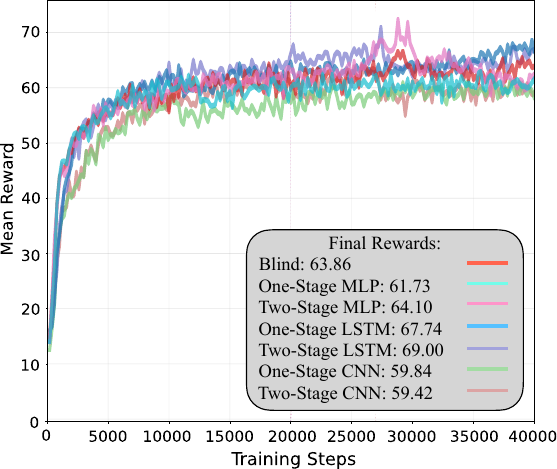}
    \caption{Training rewards from each of the policies. We see that in training, all the policies perform largely identically; we believe that the small deviations may be a function of the network architecture of that component, such as the LSTM's ability to store a hidden state or the CNN's ability to reason more spatially, but may also simply be products of randomness.}
    \label{reward_curves}
\end{figure}

\paragraph{Setup}
We train all policies in IsaacSim on a single RTX~4090 with 4096 parallel environments. Each training batch is drawn from a balanced mixture of tasks: stair ascent (25\%), stair descent (25\%), uneven terrain (25\%), and flat ground (25\%), all trained using the asymmetric actor critic algorithm \cite{pinto2017asymmetric} to 40000 steps.

\paragraph{Policies}
We compare seven variants: a blind baseline (no exteroception throughout), three one-stage perception-informed models (vision available for the full curriculum), and three two-stage models (blind in the first half of training, perception introduced in the second half). 
All actors share the same backbone sizes and differ only in network and encoder architectures: the actor is either an MLP or an LSTM with hidden layers \{512, 256, 128\}; the perception encoder is either a CNN (3$\times$3, stride~1) or an MLP, both with hidden layers \{256, 256\}. 
The critic is an MLP with the same hidden sizes and receives privileged inputs: a height scan of \(\,1.5\times 1.5\)\,m at 0.1\,m resolution, joint states, base orientation, and CoM velocities. 
Rewards and observation normalizations are held fixed across policies so that differences reflect architecture and curriculum, not reward shaping.

\begin{table}[h]
  \centering
  \small
  \caption{Environment parameters for stairs and uneven terrain.}
  \label{env_params}
  \begin{tabular}{@{}l ccc cc@{}}
    \toprule
    & \multicolumn{3}{c}{\textbf{Stairs}} & \multicolumn{2}{c}{\textbf{Uneven Terrain}} \\
    \cmidrule(lr){2-4}\cmidrule(lr){5-6}
    \textbf{Difficulty} & Height & Depth & Noise & Height & Noise \\
    \midrule
    Medium & 0.14--0.18 & 0.31 & 0.10 & 0.05--0.20 & 0.10 \\
    Hard   & 0.16--0.20 & 0.23 & 0.30 & 0.05--0.40 & 0.30 \\
    \bottomrule
  \end{tabular}
\end{table}

\begin{table}[h]
\centering
\renewcommand{\arraystretch}{1.3}
\caption{Proprioception observation terms $o_{\mathrm{name}}$ with noise, scaling, and history length applied.}
\begin{tabular}{l p{5.5cm}}
\hline
\textbf{Observation} & \textbf{Formula / Description} \\
\hline
$o_{\mathrm{base\,ang\,vel}}$ & $\omega_b \in \mathbb{R}^3$, base angular velocity in body frame. Noise $\mathcal{U}(-0.2,0.2)$, scale $0.25$. \\
$o_{\mathrm{projected\,gravity}}$ & $g_b \in \mathbb{R}^3$, gravity vector projected in body frame. Noise $\mathcal{U}(-0.05,0.05)$. \\
$o_{\mathrm{velocity\,commands}}$ & $u = (u_x,u_y,u_\omega)$, commanded base velocity, scale $(2.0,2.0,0.25)$. \\
$o_{\mathrm{joint\,pos}}$ & $q - q^{\text{default}}$, joint positions relative to defaults. Noise $\mathcal{U}(-0.01,0.01)$. \\
$o_{\mathrm{joint\,vel}}$ & $\dot q - \dot q^{\text{default}}$, joint velocities relative to defaults. Noise $\mathcal{U}(-1.5,1.5)$, scale $0.05$. \\
$o_{\mathrm{actions}}$ & $a_{k-1}$, last applied action. \\
$o_{\mathrm{phase\,obs}}$ & gait phase $\phi \in [0,1]$ with standing detection, period $0.8$. \\
\hline
\end{tabular}

\end{table}

\paragraph{Metrics}
We report:
(i) Success rate-the fraction of episodes that time out (task completed) without a fall or intervention; 
(ii) Contacts per step—the number of high-force lateral foot–environment impacts per robot step (threshold \(>\!100\)N), normalized by step count; and 
(iii) Tracking error—the mean \(\ell_2\) difference between commanded and measured base velocity, reported in cm/s.

\paragraph{Protocol}
For each policy we roll out 500 simulated units, each performing 3 episodes of 1000 steps. Stair risers are uniformly randomized and treads set per condition; uneven terrain uses block fields with specified height ranges; and uniform noise of fixed magnitude is added to the heightfield observation. Parameter ranges are summarized in Table~\ref{env_params} (units in meters).

\paragraph{Results}
On the medium (in-distribution) setting, all policies achieve near-parity; residual differences are within run-to-run variability. However, out-of-distribution (OOD) effects are more revealing:

\textit{Stairs (OOD)} All models remain reasonably strong—stairs are structured and thus easier to “overfit.” However, the two-stage variants exhibit roughly 3$\times$ lower contacts/step than their one-stage counterparts and are closer to the blind baseline in this metric. A plausible mechanism is that, under noisy exteroception, two-stage policies fall back to the robust blind stabilizer learned in stage~1, whereas one-stage policies rely more heavily on the heightfield for both planning and stabilization, leading to occasional poor foot placements. Tracking error shows a smaller but consistent improvement in the same direction.

\textit{Uneven terrain (OOD)} Here the differences are pronounced: the two-stage policies outperform one-stage by \(\approx\)10 percentage points in success on average, with corresponding reductions in contacts/step. 
Because the terrain is unstructured and harder to memorize, robustness requires both fast stabilization and longer-horizon placement—capabilities that are explicitly separated and co-trained in the two-stage pipeline but entangled in the one-stage models.

\begin{table}[h]
\centering
\renewcommand{\arraystretch}{1.3}
\caption{Nominal reward terms and weights for humanoid locomotion.}

\begin{tabular}{l l}
\hline
\textbf{Reward} & \textbf{Formula} \\
\hline
$r_{\mathrm{track\,lin\,vel\,xy\,exp}}$ & $1.0 \cdot \exp\!\left(-\tfrac{\|v^{\text{base}}_{xy}-v^{\text{command}}_{xy}\|^2}{0.25}\right)$ \\
$r_{\mathrm{track\,ang\,vel\,z\,exp}}$ & $1.0 \cdot \exp\!\left(-\tfrac{(u_{z}-\omega_{z})^2}{0.25}\right)$ \\
$r_{\mathrm{lin\,vel\,z\,l2}}$ & $-2.0 \cdot v_{z}^{2}$ \\
$r_{\mathrm{ang\,vel\,xy\,l2}}$ & $-0.05 \cdot \left(\omega_{x}^{2} + \omega_{y}^{2}\right)$ \\
$r_{\mathrm{dof\,torques\,l2}}$ & $-2.0\!\times\!10^{-5} \cdot \sum_{j} |\dot q_{j}|\,|\tau_{j}|$ \\
$r_{\mathrm{dof\,acc\,l2}}$ & $-2.5\!\times\!10^{-7} \cdot \sum_{j} \ddot q_{j}^{2}$ \\
$r_{\mathrm{dof\,vel\,l2}}$ & $-1.0\!\times\!10^{-3} \cdot \sum_{j} \dot q_{j}^{2}$ \\
$r_{\mathrm{action\,rate\,l2}}$ & $-0.01 \cdot \sum_{a} (a_t-a_{t-1})^{2}$ \\
$r_{\mathrm{undesired\,contacts}}$ & $-1.0 \cdot \sum_{b\in B}\mathbf{1}(\max_t \|F_{t,b}\|>\theta)$ \\
$r_{\mathrm{contact\,no\,vel}}$ & $-0.2 \cdot \sum_{b\in A}\|v_{b}\|^{2}\,\mathbf{1}(\text{contact}_b)$ \\
$r_{\mathrm{joint\,deviation\,hip}}$ & $-1.0 \cdot \sum_{j\in J_{\text{hip}}} |q_{j}-q_{j}^{\text{def}}|$ \\
$r_{\mathrm{joint\,deviation\,arms}}$ & $-0.5 \cdot \sum_{j\in J_{\text{arms}}} |q_{j}-q_{j}^{\text{def}}|$ \\
$r_{\mathrm{joint\,deviation\,torso}}$ & $-1.0 \cdot |q_{\text{waist}}-q_{\text{waist}}^{\text{def}}|$ \\
$r_{\mathrm{height\,torso}}$ & $-50.0 \cdot (z_{\text{root}}-0.77)^{2}$ \\
$r_{\mathrm{feet\,clearance}}$ & $+1.0 \cdot \sum_{f}\big(z_f-h_{\text{target}}(s_f)\big)^{2}\cdot(1-\text{contact}_f)$ \\
$r_{\mathrm{feet\,slide}}$ & $-0.2 \cdot \sum_{f}\|v_{f,xy}\|^{2}\,\mathbf{1}(\text{contact}_f)$ \\
$r_{\mathrm{phase\,contact}}$ & $+0.5 \cdot \sum_{f}\mathbf{1}(\text{contact}_f=\text{stance}_f)$ \\
$r_{\mathrm{stand\,still}}$ & $-0.1 \cdot \sum_{j}|q_j-q_{j}^{\text{def}}|\,\mathbf{1}(\|u\|<\epsilon)$ \\
$r_{\mathrm{feet\,flat}}$ & $-1.0 \cdot \left(1-e^{-25(|\theta^{L}_{\text{pitch}}|c_L+|\theta^{R}_{\text{pitch}}|c_R)}\right)$ \\
$r_{\mathrm{flat\,orientation\,l2}}$ & $-1.0 \cdot \left(g_x^2+g_y^2\right)$ \\
$r_{\mathrm{dof\,pos\,limits}}$ & $-5.0 \cdot \sum_{j} \text{violation}(q_j)$ \\
$r_{\mathrm{alive}}$ & $+0.15 \cdot \mathbf{1}(\neg \text{terminated})$ \\
$r_{\mathrm{termination\,penalty}}$ & $-200.0 \cdot \mathbf{1}(\text{terminated})$ \\
\hline
\end{tabular}
\end{table}

\subsection{Hardware Experiments}

To verify our hypothesis on hardware, we deploy a subset of our policies on the G1 Humanoid robot. The perception stack is run on board using the robot's Jetson Orin NX 16GB, and the RL controller is run on a Framework Laptop 13 with AMD Ryzen 7 7840u, which can either be off-board or strapped to the robot for a complete self-contained hardware stack. 
Perception data is provided by two Intel Realsense D435 cameras, one on the back of the hips, and one mounted on the chest, both pointing down. 
The lower hip camera allows for vision between the legs and behind the robot, while the upper chest camera allows for vision further in front of the robot. 
Together, they have a near-complete field of view of a 1.4m square around the robot, except for small holes where the legs shadow the camera view. Due to the edge warping, we crop the center 1m x 1m area for depth. The cameras send depth images which are merged into a combined point cloud at a rate of 30Hz.
The policy runs at 50hz, sending position setpoints to PD controllers at the joints operating at 1kHz. 

\begin{figure*}[htbp]
  \centering
 \includegraphics[width=\linewidth]{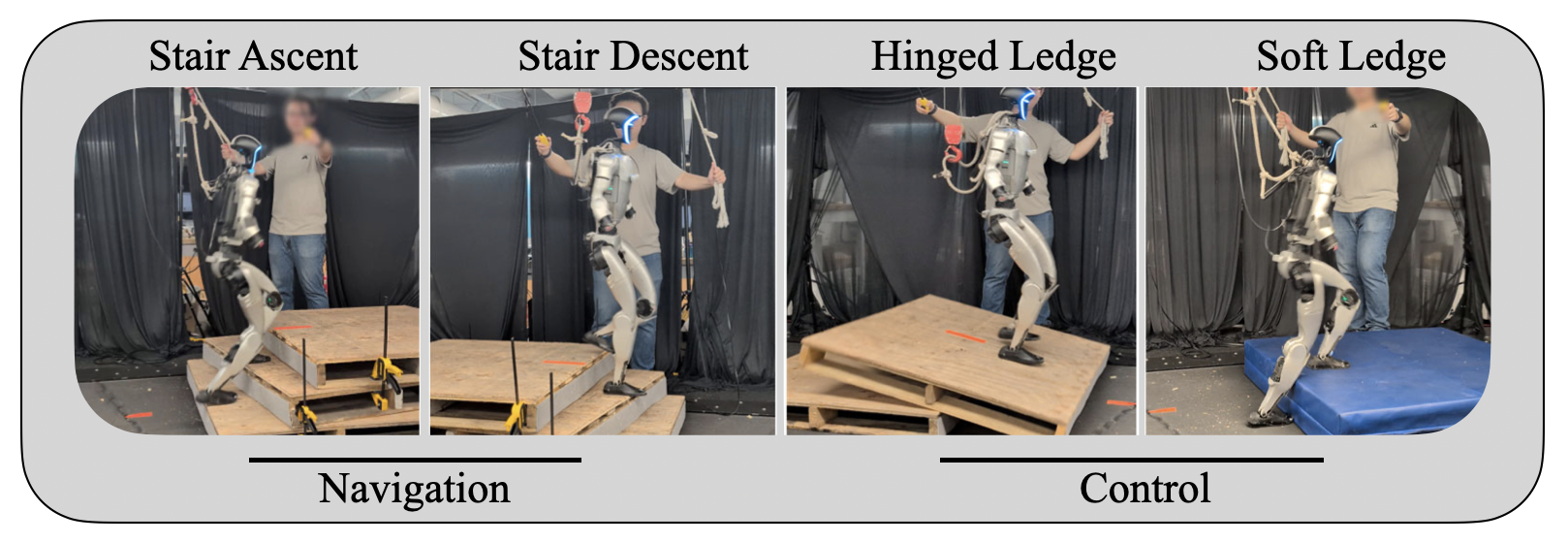}
  \caption{The four hardware experiment tasks. The first two, stair ascent and stair descent, emphasize navigation, while the second two, hinged and soft ledge, emphasize control.}
  \label{fig:task_list}
\end{figure*}
\begin{figure*}[htbp]
  \centering
 \includegraphics[width=\linewidth]{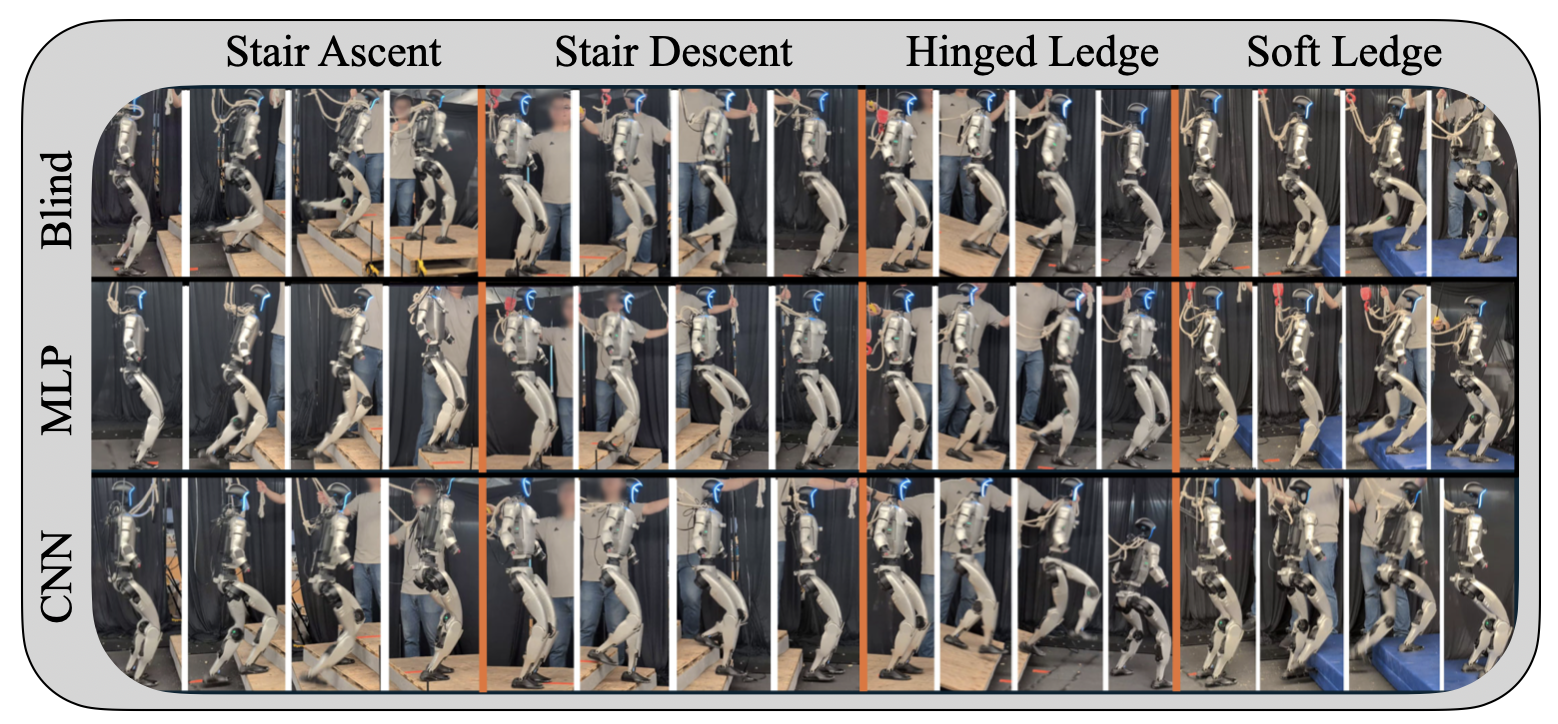}
  \caption{Blind, MLP, and CNN depicted results. The monolithic policy is not included, as for some tasks it was never successful.}
  \label{fig:tasks_run}
\end{figure*}

\begin{table*}[t]
  \centering
  \small
  \caption{Four tasks. Success rate ($\uparrow$) and tracking error in m ($\downarrow$).}
  \label{tab:tasks_all}
  \resizebox{\textwidth}{!}{%
  \begin{tabular}{@{}l cc cc cc cc@{}}
    \toprule
    & \multicolumn{2}{c}{\textbf{Stair Ascent}}
    & \multicolumn{2}{c}{\textbf{Stair Descent}}
    & \multicolumn{2}{c}{\textbf{Hinged Ledge}}
    & \multicolumn{2}{c}{\textbf{Soft Ledge}} \\
    \cmidrule(lr){2-3}\cmidrule(lr){4-5}\cmidrule(lr){6-7}\cmidrule(lr){8-9}
    \textbf{Policy}
      & $R_{\mathrm{succ}}$ (\%, $\uparrow$) & Track (m, $\downarrow$)
      & $R_{\mathrm{succ}}$ (\%, $\uparrow$) & Track (m, $\downarrow$)
      & $R_{\mathrm{succ}}$ (\%, $\uparrow$) & Track (m, $\downarrow$)
      & $R_{\mathrm{succ}}$ (\%, $\uparrow$) & Track (m, $\downarrow$) \\
    \midrule
    Blind          & 3/5 & 0.288 & 2/5 & 0.137 & 5/5 & 0.037 & 5/5 & 0.063 \\
    One-Stage MLP  & 1/5 & 0.000 & 1/5 & 0.000 & 0/5   & --    & 0/5   & --    \\
    \bf{Two-Stage CNN}  & \bf{4/5} & \bf{0.175} & \bf{5/5} & \bf{0.228} & \bf{5/5} & \bf{0.059} & \bf{5/5} & \bf{0.167} \\
    \bf{Two-Stage MLP}  & \bf{4/5} & \bf{0.045} & \bf{5/5} & \bf{0.163} & \bf{5/5} & \bf{0.199} & \bf{4/5}  & \bf{0.113} \\
    \bottomrule
  \end{tabular}
  }
\end{table*}

\paragraph{Hardware tasks}
We evaluate on four hardware tasks designed to probe different components of the stack: stair ascent, stair descent, hinged ledge, and soft ledge. 
The stair tasks comprise a short flight of three steps (riser $\approx\!18$\,cm) with small landings ($\sim\!20$\,cm) and a horizontal skew of $\sim\!25^\circ$. This geometry forces careful toe/heel placement and weight transfer—missteps induce lateral perturbations—so these tasks primarily stress navigation (longer-horizon footstep/velocity planning). 
The ledge tasks use a 36\,cm elevation change with transient compliance: the hinged variant is a plank balanced on a pivot that tips under load, and the soft variant lands onto a compliant gym mat. Perception sees a nominal ledge, but the dominant difficulty is the unmodeled, state-dependent disturbance at contact; these tasks primarily stress control (fast stabilization under transients).

\paragraph{Policies and trials}
For each task we run five trials for each of four policies:
(i) a blind baseline, 
(ii) a one-stage MLP,
(iii) a two-stage MLP, and 
(iv) a two-stage CNN+MLP.

\paragraph{Metrics}
We report two task-level metrics. Success rate counts trials that complete the task without a fall or human intervention.
Precision measures repeatability: from a standing start we command $0.3$\,m/s forward, stop after task completion, and record the net lateral drift. We report the mean absolute deviation across the five trials to suppress fixed biases and emphasize stability and consistency.

\paragraph{Observations}
First, the one-stage MLP underperforms across terrains despite identical training conditions. Empirically, footstep selection often appears reasonable, but stabilization degrades quickly, consistent with over-reliance on noisy heightfields for low-level control.
Second, the blind policy transfers reasonably well and is particularly strong on the ledge (control-dominant) tasks, but fails more frequently on stairs where precise edge-aware placement is required (missed steps when descending; edge strikes when ascending).
Finally, the two-stage policies (MLP and CNN+MLP) perform similarly and robustly on both navigation- and control-dominant tasks, supporting our central claim: once the layered structure is in place, the specific encoder and backbone choice is of lesser importance. For absolute peak performance, one could further tune architectures or add specialized modules, but our results indicate such complexity is unnecessary to achieve robust behavior on these tasks.

\bibliographystyle{IEEEtran}
\bibliography{refs}
\end{document}